\pdfoutput=1

\documentclass[5p,times]{elsarticle}

\makeatletter
\def\ps@pprintTitle{%
	\let\@oddhead\@empty
	\let\@evenhead\@empty
	\let\@oddfoot\@empty
	\let\@evenfoot\@oddfoot
}
\makeatother

\usepackage{lineno,hyperref}
\modulolinenumbers[5]

\usepackage{amssymb}
\usepackage{amsmath}
\usepackage{tabularx}
\usepackage{multirow}
\usepackage{placeins}

\usepackage{framed}
\usepackage{fancyvrb}
\usepackage{algpseudocode}
\usepackage{algorithm}

\usepackage[utf8]{inputenc}
\usepackage[T1]{fontenc}
\usepackage{natbib}

\DeclareMathOperator{\atan2}{atan2}

\usepackage[font=small,skip=5pt]{caption}

\usepackage{hyperref} 

\hypersetup{
	unicode=false,          
	pdftoolbar=true,        
	pdfmenubar=true,        
	pdffitwindow=false,     
	pdfstartview={FitH},    
	pdfnewwindow=true,      
	colorlinks=true,       
	linkcolor=oceanboatblue,          
	citecolor=napiergreen,  
	filecolor=magenta,      
	urlcolor=cyan           
}

\begin{document}

\begin{frontmatter}

\title{Anakatabatic Inertia: Particle-wise Adaptive Inertia for PSO}


\author[mymainaddress]{Siniša Družeta \corref{mycorrespondingauthor}}
\cortext[mycorrespondingauthor]{Corresponding author}
\ead{sinisa.druzeta@riteh.hr}

\author[mymainaddress]{Stefan Ivić}



\address[mymainaddress]{University of Rijeka Faculty of Engineering, Vukovarska 58, Rijeka, Croatia}

\begin{abstract}
Throughout the course of the development of Particle Swarm Optimization, particle inertia has been established as an important aspect of the method for researching possible method improvements. As a continuation of  our previous research, we propose a novel generalized technique of inertia weight adaptation based on individual particle's fitness improvement, called anakatabatic inertia. This technique allows for adapting inertia weight value for each particle corresponding to the particle's increasing or decreasing fitness, i.e. conditioned by particle's ascending (anabatic) or descending (katabatic) movement. The proposed inertia weight control framework was metaoptimized and tested on the 30 test functions of the CEC 2014 test suite. The conducted procedure produced four anakatabatic models, two for each of the PSO methods used (Standard PSO and TVAC-PSO). The benchmark testing results show that using the proposed anakatabatic inertia models reliably yield moderate improvements in accuracy of Standard PSO (final fitness minimum reduced up to 0.09 orders of magnitude) and rather strong improvements for TVAC-PSO (final fitness minimum reduced up to 0.59 orders of magnitude), mostly without any adverse effects on the method's performance.
\end{abstract}


\end{frontmatter}
\section{Introduction}
\label{sec:Introduction}
Particle Swarm Optimization (PSO) is an optimization method originally inspired by the movement of bird flocks and fish schools \cite{PSO1,PSO2}. The method tracks a group of agents (called particles) moving through the search space, each agent adapting its movement on the basis of its own findings as well as the findings of other agents. To this day, a great number of  modifications and improvements have been proposed for PSO, as both elegance and capability of the method keep motivating researchers to further investigate its features and advance its performance.

In an attempt to enhance the effectiveness of PSO by enabling the particles with a certian ``awareness'' of their own improvement, we have previously proposed an enhancement of PSO method with Personal Fitness Improvement Dependent Inertia (PFIDI) which makes each particle's movement conditioned by its improvement in fitness. The proposed PFIDI technique of moving particles through the search space is called Languid Particle Dynamics (LPD) and the corresponding LPSO variant is called Languid PSO (LPSO)\cite{LPSO,LPSOegyptian}. Tested on a spectrum of several PSO variants, method parameter configurations, and goal functions, the LPD technique was shown to predominantly produce an increase in PSO accuracy and stability.

In this paper we present a novel, generalized PFIDI technique, which allows for finer fitness improvement based inertia adaptation than it was proposed in the rather blunt LPD approach.

\section{PSO and Languid Particle Dynamics}
\label{sec:PSOandLPD}

In standard PSO the particles move with a certain amount of inertia through the search space, while being attracted to the best position that they individually have found, and to the best position found by any particle of their neighborhood.

For each individual particle of the PSO swarm, we keep track of its position in the $D$-dimensional search space $\mathbf{x} = (x_1, x_2, x_3, ... x_D)$, its historically best position $\mathbf{p}$, its current velocity $\mathbf{v}$ and historically best position of its neighboring particles $\mathbf{g}$. After random initialization of positions $\mathbf{x}$ and velocities $\mathbf{v}$, a $k$-th particle moves by updating its velocity and position at iteration $t$ \cite{PSO98}:
\begin{equation}
\begin{aligned}
\label{eq:PSOv}
\mathbf{v}_k^{(t)} ={} &  w_k^{(t)} \cdot \mathbf{v}_k^{(t-1)} + c_1\cdot \mathbf{r}_1\circ\left( \mathbf{p}_k^{(t-1)}-\mathbf{x}_k^{(t-1)}\right) \\  
& + c_2\cdot \mathbf{r}_2\circ\left( \mathbf{g}_k^{(t-1)}-\mathbf{x}_k^{(t-1)}\right) ~,
\end{aligned}
\end{equation}
\begin{equation}
\label{eq:PSOx}
\mathbf{x}_k^{(t)} = \mathbf{x}_k^{(t-1)} + \mathbf{v}_k^{(t)}~,
\end{equation}
where  $w$ is the inertia weight factor, $c_1$ and $c_2$ are cognitive and social PSO coefficients, respectively, while $\mathbf{r}_1$ and $\mathbf{r}_2$ are  $D$-dimensional vectors of random numbers in the range $[0,1]$. Note that vector multiplication in \eqref{eq:PSOv} is a Hadamard product. The swarm consists of $n$ particles, i.e. $k = 1...n$.

Since different optimization problems require different convergence dynamics, many various methods have been proposed for dynamically changing or adapting inertia weight \cite{IWSTRAT}. Still, inertia weight is often used as a constant, when it is generally recommended to use $w = 0.7 \pm 0.05$ \cite{PHDTHESIS,STANDARDPSO}.

Coefficients $c_1$ and $c_2$ are traditionally used as $c_1 = c_2 = 2.0$, but better understanding of their influence encourages the use of lower values and problem-specific calibration \cite{PHDTHESIS,STANDARDPSO}, as well as changing them over iterations \cite{TVAC}.

The version of PSO with each particle being informed about best found locations by the entire swarm is called ``\emph{gbest} PSO'', while the version with each particle communicating only with a subset of the swarm is called ``\emph{lbest} PSO'' \cite{PSO2}. In other words, a swarm may be entirely connected in a single neighborhood or divided in many smaller neighborhoods, with neighborhood topologies being purely index-based, i.e. not related to search space locality. Many different options have been proposed for the neighborhood topology of the \emph{lbest} PSO and no specific topology has been universally adopted as most beneficial in terms of overall PSO performance. Standard PSO implementations mostly imply the use of simple circular (``ring'') topology \cite{STANDARDPSO} or random topology \cite{PSOCODE}.

Note that, since standard PSO particles do not track their fitness progress, they have no information on the fitness change along their path. In the course of PSO research, a number of techniques for using fitness improvement information for improving the efficiency of the swarming process have been developed, albeit only a few of those consider particle's own fitness improvement in each particle's movement logic. On the other hand, although some authors have proposed particle-wise inertia control, in standard PSO as well as in most other PSO variants inertia weight $w$ is considered to be global (i.e. one and the same for the entire swarm). 

Based on utilizing particle-wise inertia control for resolving personal fitness improvement dependent particle movement, the PFIDI approach and the LPSO method \cite{LPSO,LPSOegyptian} were proposed in which each particle tracks its fitness evolution so that this information can be used for altering its movement process. This information is then used in a switch-like condition on inertia term of each individual particle:
\begin{equation}
\label{eq:wLPSO}
w_k^{(t)} =
\begin{cases}
\left(w^{(0)}+0.05\right)  & \text{when $f\left( \mathbf{x}_k^{(t-1)}\right)  < f\left( \mathbf{x}_k^{(t-2)}\right) $} \\
0 &\text{otherwise}
\end{cases}~,
\end{equation}
where $w^{(0)}$ is the initial (``default'') value of inertia weight and $f$ is the fitness function. This means that the $k$-th particle has inertia only as long as it keeps advancing in a direction of better fitness (the formulation \eqref{eq:wLPSO} assumes a minimization problem). The correction of +0.05 for inertia weight when inertia is not disabled was proposed in \cite{LPSOegyptian} so as to compensate for the reduced overall velocity of the swarm due to the intermittent inertial velocity of the particles.

Behaving in this manner, a particle disregards its previous direction if it failed to take it to a better location. Since such particle behavior implies a certain lack of enthusiasm, the adjective ``languid'' was adopted as a designator of this type of particle movement dynamics. 

PFIDI techniques has also been successfully implemented in \cite{LPSOegyptian} in Time Varying Acceleration Coefficients PSO (TVAC-PSO) \cite{TVAC}, a PSO variant which uses linearly changing coefficients $c_1$ and $c_2$, contrary to the standard PSO where coefficients $c_1$ and $c_2$ are used as constant values. Coefficients $c_1$ and $c_2$ calculated by:
\begin{equation}
\label{eq:TVACc1}
c_1^{(t)} = c_{1_s} + \left(c_{1_f}-c_{1_s}\right) \frac{t}{t_{max}}
\end{equation}
\begin{equation}
\label{eq:TVACc2}
c_2^{(t)}  = c_{2_s} + \left(c_{2_f}-c_{2_s}\right) \frac{t}{t_{max}}
\end{equation}
where $c_{1_s}$, $c_{1_f}$, $c_{2_s}$ and $c_{2_f}$ represent starting and final values of coefficients $c_1$ and $c_2$ which are linearly increasing/decreasing over iterations of the PSO swarming process. In the TVAC-PSO implementation used for this research $c_1$ is decreasing from $c_{1_s}=2.5$ to $c_{1_f}=0.5$ while $c_2$ is increasing from $c_{2_s}=0.5$ to $c_{2_f}=2.5$, as recommended by the authors of the method. 

TVAC-PSO also features Linearly Decrasing Inertia Weight (LDIW) \cite{LDIW}:
\begin{equation}
\label{eq:wLDIW}
w^{(t)} = w_{min} + \left(w_{max}-w_{min}\right) \frac{t}{t_{max}} ~,
\end{equation}
which is a fairly popular PSO inertia handling technique. Here $w_{min}$ is the minimum value of interia weight factor, $w_{max}$ is the maximum value of intertia weight factor, $t$ is the current iteration for which $w$ is calculated, and $t_{max}$ is the maximum number of allowed iterations (corresponding to maximum allowed function evaluations $eval_{max}$).

Both standard PSO and TVAC-PSO were used for the implementation of the novel adaptive inertia technique proposed in this paper.

\section{Anakatabatic inertia}
\label{sec:Anakatabatic}

As a generalization of LPD which would take into account a particle's advancement as well as search progress of the entire swarm, here we propose a novel particle-wise fitness based inertia weight adaptation technique. Considering that it employs adapting inertia weight values corresponding to the particle's ascending (anabatic) or descending (katabatic) movement (in terms of increasing or decreasing fitness), the authors adopted the name ``anakatabatic inertia'' for the proposed technique.

As a basis of the anakatabatic inertia scheme, we introduce the $\theta$ parameter, defined as:
\begin{align}
\label{eq:theta}
\theta_k^{(t)} = \atan2 \left( \Delta f\left( \mathbf{x}_k^{(t)}\right), \min \Delta f\left( \mathbf{x}_i^{(t)}\right) \right), ~i = 1...n ~,
\end{align}
where $\Delta f\left( \mathbf{x}_k^{(t)}\right) $ is the fitness change of the $k$-th particle since previous iteration, i.e. $\Delta f\left( \mathbf{x}_k^{(t)}\right)  = f\left( \mathbf{x}_k^{(t)}\right)  - f\left( \mathbf{x}_k^{(t-1)}\right) $ (where less is better, assuming minimization). It should be noted that the used $\atan2$ implementation yields only positive values of $\theta$, thus for $\Delta f\left( \mathbf{x}_k^{(t)}\right) < 0$ (third quadrant), $\theta \in \left[\pi, \frac{5\pi}{4}\right]$. Also, since $\atan2$ of near zero arguments returns $\theta \approx 0$, the values of $\theta < 10^{-300}$ are replaced with random values taken from $\left[\frac{\pi}{4}, \frac{5\pi}{4}\right]$.

\begin{figure}
	\centering
	\includegraphics[width=0.9\linewidth]{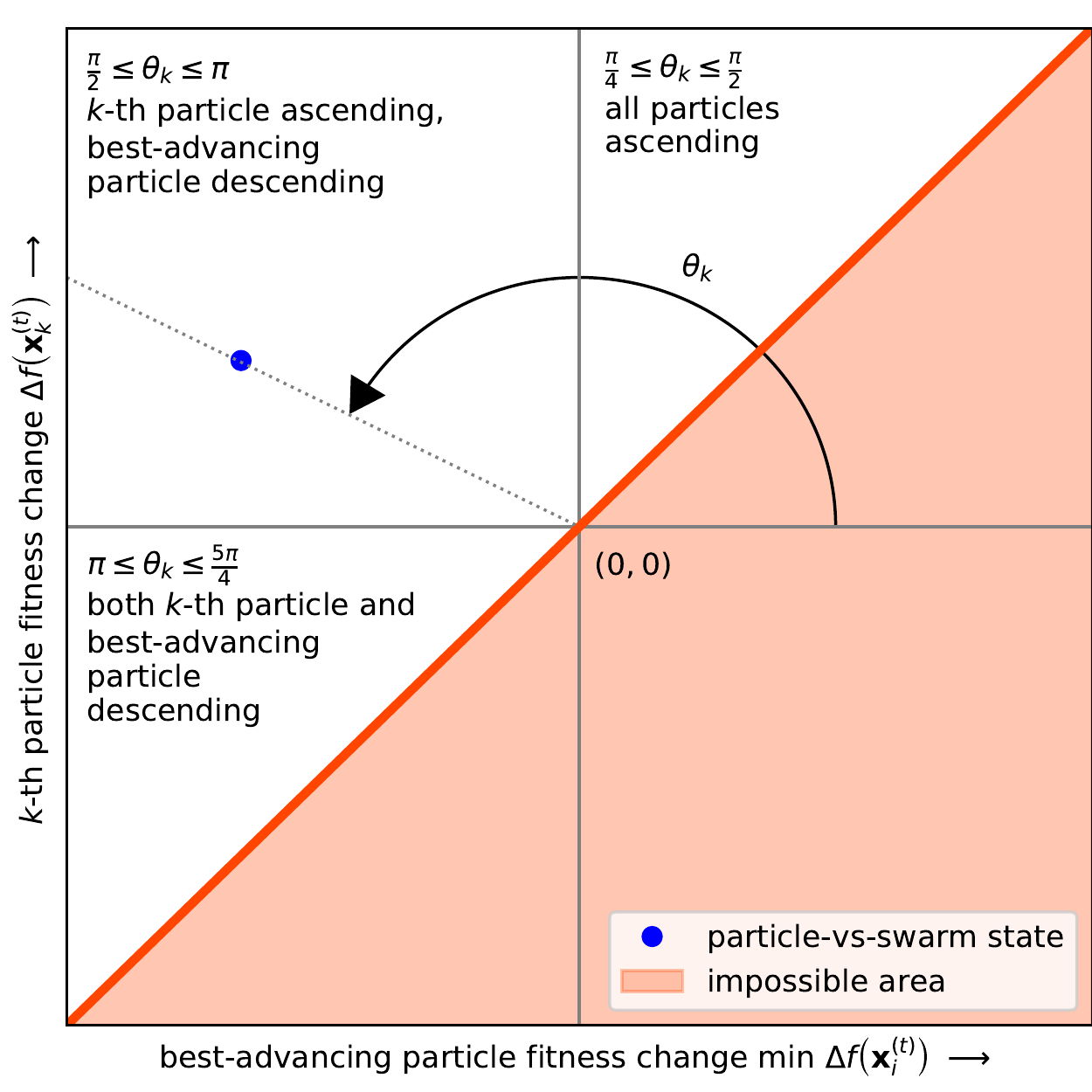}
	\caption{Assessing particle-versus-swarm advancement state by use of $\theta$ parameter (assuming minimization)}
	\label{fig:akbTheta}
\end{figure}

Defined in this way, $\theta$ parameter allows for tracking three distinct particle-vs-swarm advancement states: for $\frac{\pi}{4} \leq \theta_k \leq \frac{\pi}{2}$ all particles (including the $k$-th particle) have failed in improving their fitness, for $\frac{\pi}{2} \leq \theta_k \leq \pi$ some of the particles have improved their fitness, but the $k$-th particle has not, and for $\pi \leq \theta_k \leq \frac{5\pi}{4}$ some of the particles, including the $k$-th particle, have managed to improve their fitness. This is visually explained in Figure \ref{fig:akbTheta}.

Now we can use $\theta_k^{(t)}$ values to obtain inertia weight $w_k^{(t)}$ values. For this we introduce a function $W(\theta)$, so that we can compute $w_k^{(t)} = W\left(\theta_k^{(t)}\right)$. Following up on the idea of linearly changing parameters of TVAC and LDIW, we will define $W(\theta)$ as a linear combination of two interpolation functions:
\begin{equation}
\label{eq:W}
W\left(\theta_k^{(t)}\right) = W_s\left(\theta_k^{(t)}\right) + \left( W_f\left(\theta_k^{(t)}\right) - W_s\left(\theta_k^{(t)}\right) \right) \frac{t}{t_{max}} ~,
\end{equation}
where $W_s(\theta)$ and $W_f(\theta)$ produce starting and final values of inertia weight.

The pseudo-code of standard PSO enhanced with anakatabatic inertia is given in Algorithm 1.

\begin{algorithm}[H]
	\caption{Standard PSO with anakatabatic inertia}
	\begin{algorithmic}[1]
		\For{particle $k = 1$ \textbf{to} $n$}
		\State number of evaluations $eval=0$
		\State iteration $t=0$
		\State initialize particle position $\mathbf{x}_k^{(t)}$ and velocity $\mathbf{v}_k^{(t)}$
		\State evaluate fitness $f(\mathbf{x}_k^{(t)})$, $eval = eval + 1$
		\State find personal and neighborhood best: $\mathbf{p}_k^{(t)}$, $\mathbf{g}_k^{(t)}$
		\EndFor
		\While{$eval < eval_{max}$}
		\State iteration $t=t+1$
		\For{particle $k = 1$ \textbf{to} $n$}
		\If{$t \ge 2$}
		\State calculate theta $\theta_k^{(t)}$ \Comment{Eq. \eqref{eq:theta}}
		\State calculate inertia weight $w_k^{(t)} = W\left(\theta_k^{(t)}\right)$ \Comment{Eq. \eqref{eq:W}}
		\EndIf
		\State calculate new velocity $\mathbf{v}_k^{(t)}$ \Comment{Eq. \eqref{eq:PSOv}}
		\State calculate new position $\mathbf{x}_k^{(t)}$ \Comment{Eq. \eqref{eq:PSOx}}
		\State evaluate fitness $f(\mathbf{x}_k^{(t)})$, $eval = eval + 1$
		\State find personal and neighborhood best: $\mathbf{p}_k^{(t)}$, $\mathbf{g}_k^{(t)}$
		\EndFor
		\EndWhile
	\end{algorithmic}
\end{algorithm}

\section{Metaoptimization of anakatabatic models}
\label{sec:Metaoptimization}

By use of $W_s(\theta)$ and $W_f(\theta)$ functions we can define a wide variety of PFIDI schemes. For example, the LPD technique given in \eqref{eq:wLPSO} can be redefined as:
\begin{equation}
\label{eq:akbLPSO}
W_s(\theta) = W_f(\theta) =
\begin{cases}
0  & \text{for $\frac{\pi}{4} \leq \theta \leq \pi$ } \\
\left(w^{(0)}+0.05\right) & \text{for $\pi \leq \theta \leq \frac{5\pi}{4}$ }
\end{cases}~,
\end{equation}

To keep things pragmatic and practical, for this research we will define $W_s(\theta)$ and $W_f(\theta)$ as simple linear interpolation functions based on five values of $w$ at five specific values of $\theta$, namely $\theta \in \left\lbrace \frac{\pi}{4}, \frac{\pi}{2}, \frac{3\pi}{4}, \pi, \frac{5\pi}{4}\right\rbrace $. This allows for searching for efficient $W_s(\theta)$ and $W_f(\theta)$ by metaoptimizing these ten values of inertia weight (five for $W_s(\theta)$ and five for $W_f(\theta)$). The $W_s(\theta)$, $W_f(\theta)$ function pair make for what we will call an ``anakatabatic model'', which will be defined by the ten $w$-values, forming the metaoptimization design vector:
\begin{align}
\label{eq:akbX}
\mathbf{X}_M = &\left( 
W_s\left(\frac{\pi}{4}\right), W_s\left(\frac{\pi}{2}\right), W_s\left(\frac{3\pi}{4}\right), W_s\left(\pi\right), W_s\left(\frac{5\pi}{4}\right),\right. \nonumber \\
& \left. W_f\left(\frac{\pi}{4}\right), W_f\left(\frac{\pi}{2}\right), W_f\left(\frac{3\pi}{4}\right), W_f\left(\pi\right), W_f\left(\frac{5\pi}{4}\right) 
\right)~.
\end{align}

For the metaoptimization fitness we used CEC 2014 test results. CEC 2014 test was designed for benchmarking of real-parameter single objective optimization algorithms and comprises 30 test functions, most of which have randomly shifted global optima, while all are randomly rotated (see \cite{CEC2014} for details). The test consists of 3 unimodal functions (F1, F2, and F3), 13 shifted multimodal functions (F4, F5, ..., F16), 6 hybrid functions based on unimodal functions and shifted multimodal functions (F17, F18, ..., F22), and 8 composition functions based on unimodal functions, shifted multimodal functions and hybrid functions (F23, F24, ..., F30). Although CEC 2014 test functions support four search space dimensionalities $D \in \{ 10, 20, 50, 100 \}$, for metaoptimization either the test functions of $D = 10$ or the test functions of $D = 20$ were used.

As for the PSO parameters, $w^{(0)} = 0.72$ was used, as well as $c_1 = c_2 = 1.0$ in standard PSO. Only the \emph{gbest} version of PSO was used. The number of PSO particles was kept at $3D$.

The metaoptimization fitness function was obtained by use of  best-of-swarm fitness errors $\varepsilon$, computed with $10^{3}D$ function evaluations on each CEC 2014 test function:
\begin{equation}
\label{eq:epsilonmean}
\varepsilon=f_{best}-f^\star~,
\end{equation}
where $f_{best}$ stands for final best-of-swarm fitness value, averaged across 250 or 500 computational runs, while $f^\star$ stands for known global minimum of goal function $f$. These $\varepsilon$ values were then sorted by value and used for computing metaoptimization fitness $F_M$ as follows:
\begin{equation}
\label{eq:metaoptf}
F_M \left( \mathbf{X}_M \right) = \frac{1}{24} \sum_{\underset{\text{(sorted)}}{i=4}}^{27} \log \varepsilon_i~,
\end{equation}
where $\varepsilon_i$ stands for the fitness error of the $i$-th CEC 2014 test function. By using only the middle 80\% of the test functions (after sorting by $\varepsilon$, only functions $4...27$ are included in the summation) some possible outliers (i.e. extremely well performing and extremely poorly performing test functions) are being excluded from the metaoptimization fitness, i.e. overspecialization has hopefully been avoided.

The bounds for the members of the design vector $\mathbf{X}_M$ \eqref{eq:akbX} were defined as $\left[-2, 2\right]$. Note that this allows for a wide range of $w$-values and also consequently allows for employing negative inertia if beneficial. Both Standard and TVAC-PSO were used for metaoptimization.

Due to the high computational cost of this metaoptimization process, the computations were performed on the BURA supercomputer of the University of Rijeka Center for Advanced Computing and Modeling.

Through multiple conducted metaoptimization runs, accompanied by some manual trial and error experimentation, several anakatabatic models were found. Some of the models were found by iterative metaoptimization in which previously found models were used in anakatabatic-PSO based metaoptimization, through which subsequent (``next generation'') models were obtained. 

\begin{figure}
	\includegraphics[width=\linewidth]{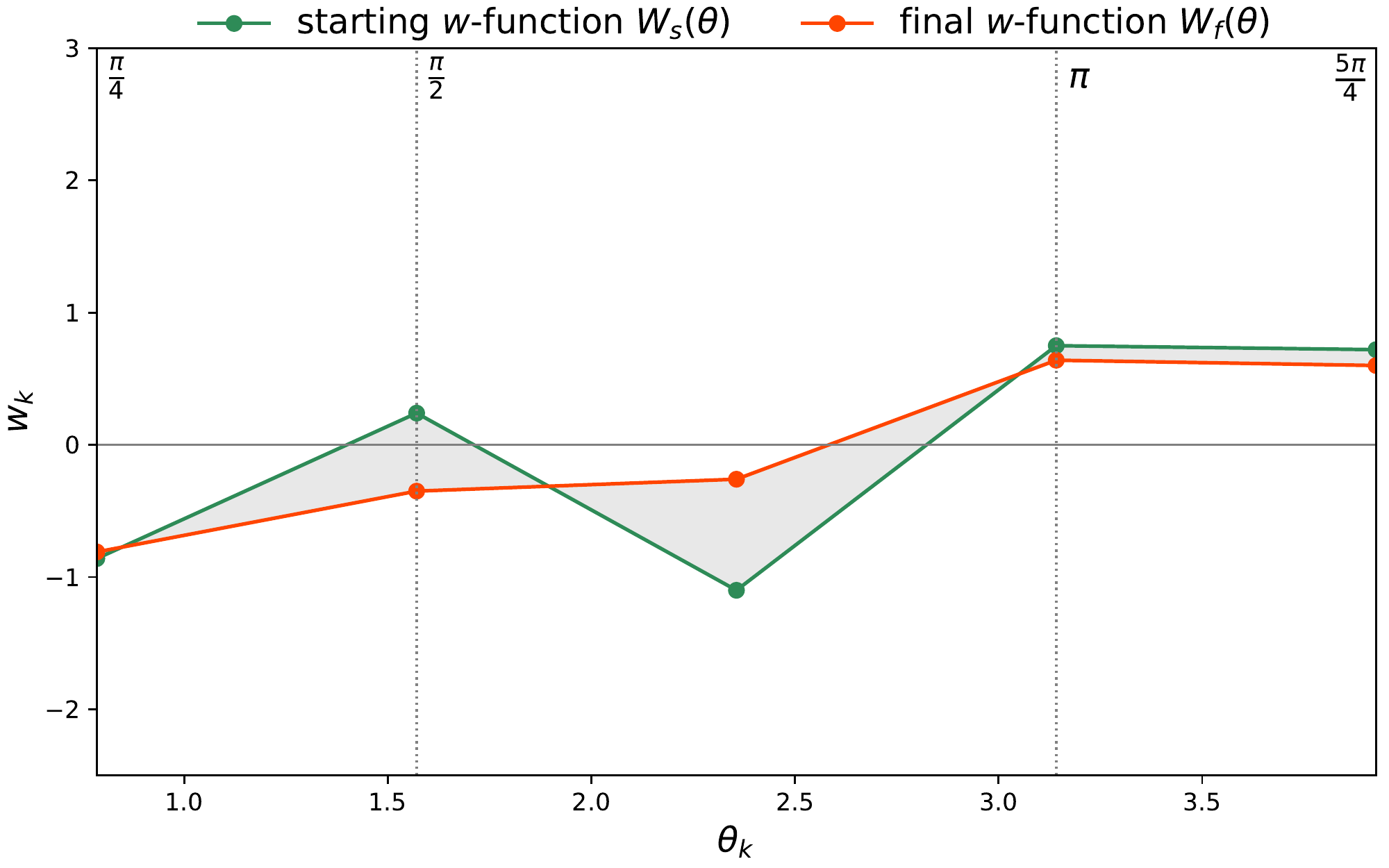}
	\caption{Standard PSO anakatabatic model ``Flying Stork''}
	\label{fig:akbFlyingStork}
\end{figure}

\begin{figure}
	\includegraphics[width=\linewidth]{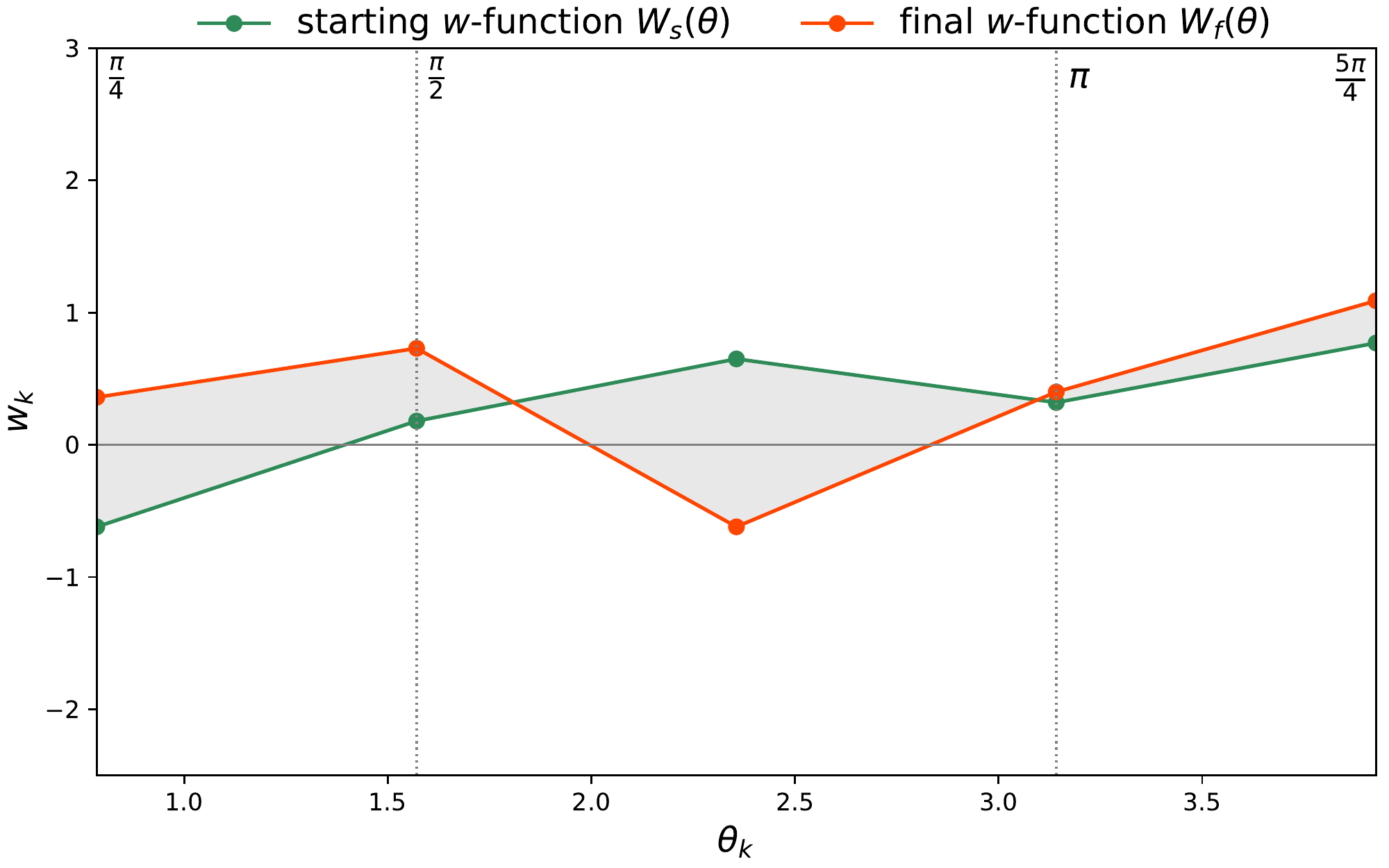}
	\caption{Standard PSO anakatabatic model ``Messy Tie''}
	\label{fig:akbMessyTie}
\end{figure}

\begin{figure}
	\includegraphics[width=\linewidth]{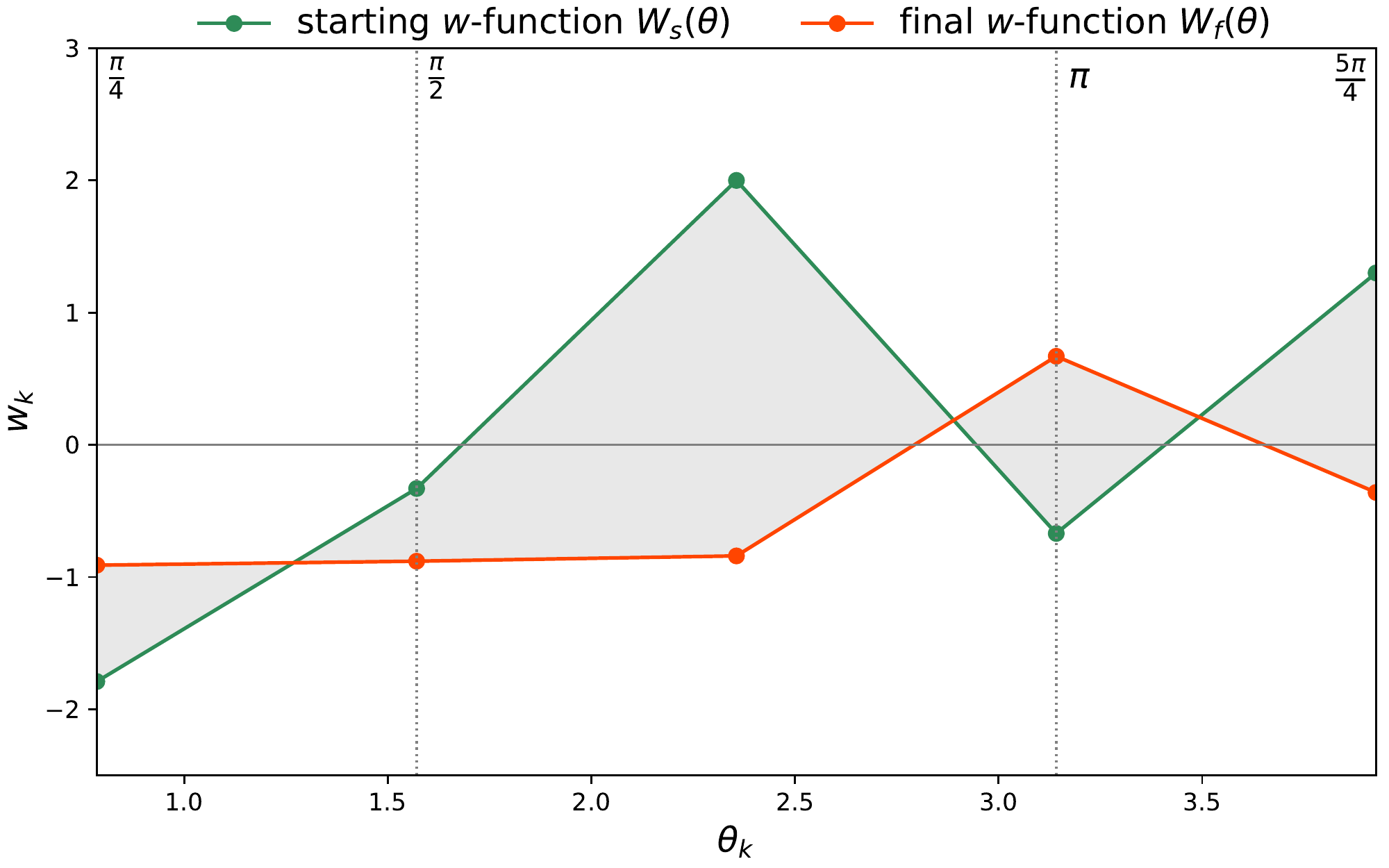}
	\caption{TVAC-PSO anakatabatic model ``Rightward Peaks''}
	\label{fig:akbRightwardPeaks}
\end{figure}

\begin{figure}
	\includegraphics[width=\linewidth]{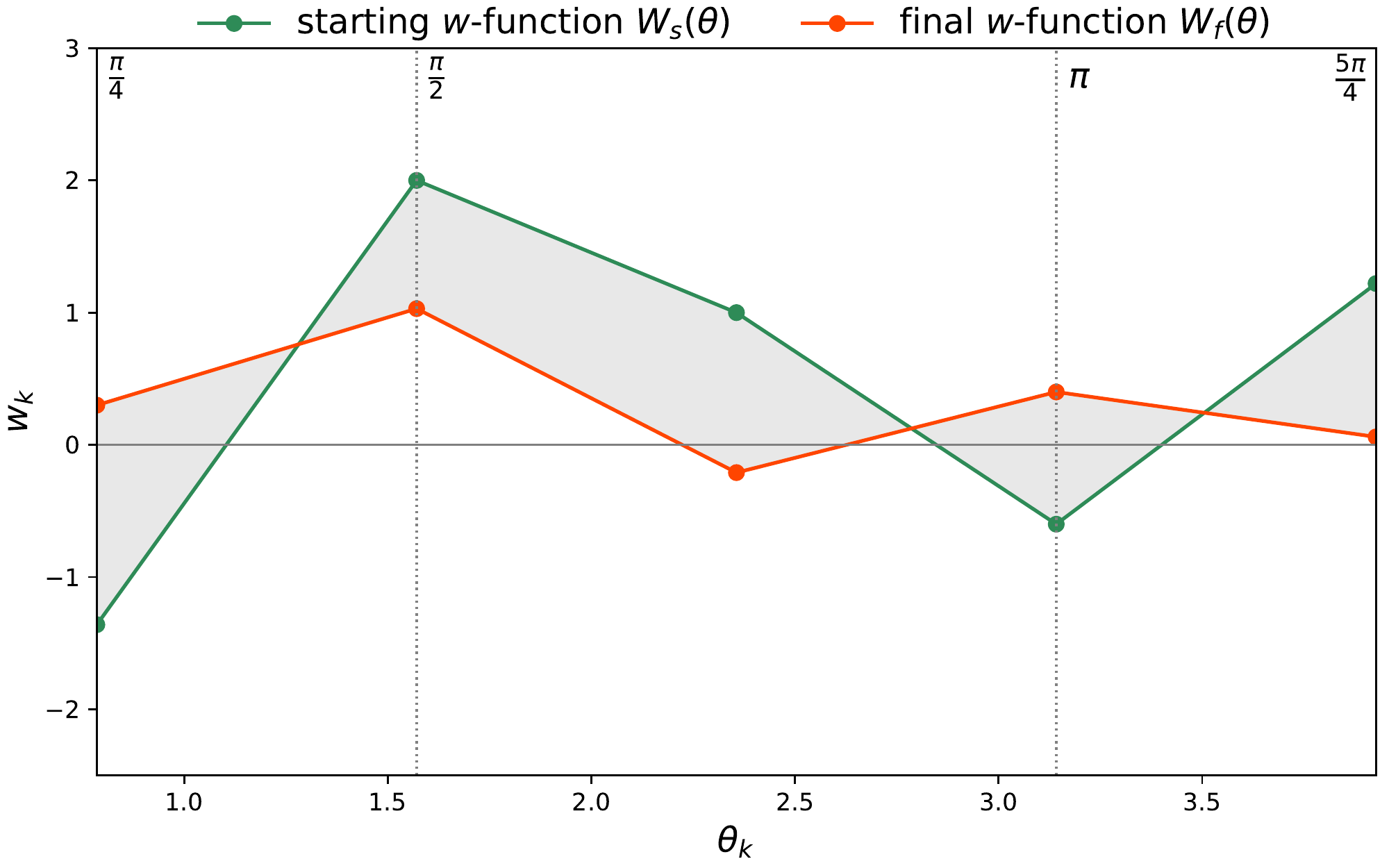}
	\caption{TVAC-PSO anakatabatic model ``Origami Snake''}
	\label{fig:akbOrigamiSnake}
\end{figure}

Overall best-performing anakatabatic models are given in Table \ref{tab:AkbModels}, and also graphically shown in Figures \ref{fig:akbFlyingStork}-\ref{fig:akbOrigamiSnake}. Just these visual representations of the anakatabatic models themselves indicate that the metaoptimization problem is very hard and strongly multimodal.

\begin{table*}[!]\footnotesize
	\caption{Anakatabatic models}
	\label{tab:AkbModels}
	\begin{tabular*}{\textwidth}{l@{\extracolsep{\fill}}cccccc}
		\hline
		\multicolumn{1}{l}{\multirow{2}{*}{PSO variant}}  &
		\multicolumn{1}{l}{\multirow{2}{*}{Anakatabatic model}}  & \multicolumn{5}{c}{$W_s(\theta)$, $W_f(\theta)$} \\
		\multicolumn{1}{c}{} & \multicolumn{1}{c}{} & \multicolumn{1}{c}{$\frac{\pi}{4}$} & \multicolumn{1}{c}{$\frac{\pi}{2}$}  & \multicolumn{1}{c}{$\frac{3\pi}{4}$} & \multicolumn{1}{c}{$\pi$} & \multicolumn{1}{c}{$\frac{5\pi}{4}$}  \\
		\hline
		Standard PSO & ``Flying Stork'' (Fig. \ref{fig:akbFlyingStork}) & -0.86, -0.81 & 0.24, -0.35 & -1.10, -0.26 & 0.75, 0.64 & 0.72, 0.60 \\
		Standard PSO & ``Messy Tie'' (Fig. \ref{fig:akbMessyTie}) & -0.62, 0.36 & 0.18, 0.73 & 0.65, -0.62 & 0.32, 0.40 & 0.77, 1.09 \\
		TVAC-PSO & ``Rightward Peaks'' (Fig. \ref{fig:akbRightwardPeaks}) & -1.79, -0.91 & -0.33, -0.88 & 2.00, -0.84 & -0.67, 0.67 & 1.30, -0.36 \\
		TVAC-PSO & ``Origami Snake'' (Fig. \ref{fig:akbOrigamiSnake}) & -1.36, 0.30 & 2.00, 1.03 & 1.00, -0.21 & -0.60, 0.40 & 1.22, 0.06 \\
		\hline
	\end{tabular*}
\end{table*}

\section{Benchmark testing anakatabatic models}
\label{sec:AkbTesting}

The obtained anakatabatic models were tested in more detail on the CEC 2014 test. For comparing the accuracy of the two used PSO variants with the accuracy of their respective sub-variants enabled with anakatabatic inertia, best-of-swarm fitness errors were used, computed with $10^{3}D$ function evaluations:
\begin{equation}
\label{eq:epsilonmeanhat}
\hat{\varepsilon} = \hat{f}_{best} - f^\star~.
\end{equation}
Here $\hat{f}_{best}$ stands for final best-of-swarm fitness value averaged across 1000 computational runs, while $f^\star$ stands for known global minimum of goal function $f$. It should be noted that throughout the entire benchmark testing no occurrences of   $\hat{\varepsilon} = 0$ were found.

Furthermore, so as to provide a relative comparison of the selected methods' performance, a dimensionless rating $\alpha$ is used \cite{LPSO}: 
\begin{equation}
\label{eq:alpha}
\alpha = \frac{\hat{\varepsilon}_{X}-\hat{\varepsilon}_{X_A}}{\frac{1}{2}(\hat{\varepsilon}_{X}+\hat{\varepsilon}_{X_A})}~,
\end{equation}
where $\varepsilon_{X}$ and $\varepsilon_{X_L}$ represent $\varepsilon$ values for a specific pure (``$X$'') PSO variant (Standard PSO, TVAC-PSO) and its corresponding counterpart (``$X_A$'') with anakatabatic inertia enabled (Standard PSO with anakatabatic inertia, TVAC-PSO with anakatabatic inertia). A measure of this kind is easy to understand ($\alpha>0$ means that anakatabatic variant performed better than pure variant and vice versa) and may reasonably be averaged across test functions and then used as a bulk value representing overall method score, with values confined to the interval $[-2, 2]$. 

Additionally, one another measure of method success, similar to the metaoptimization fitness function \eqref{eq:metaoptf}, was used for benchmark testing:
\begin{equation}
\label{eq:omega}
\Omega = \log \frac{\hat{\varepsilon}_{X}}{\hat{\varepsilon}_{X_A}}~,
\end{equation}
representing the average improvement of the fitness error order of magnitude when comparing anakatabatic PSO variant $X_A$ with its pure counterpart $X$ (where $\Omega>0$ means that anakatabatic variant performed better than pure variant and vice versa). A measure of this kind balances out the wide range of error orders of magnitude for the 30 test functions.

The results of the testing of the effects of anakatabatic inertia on the two PSO variants for $D \in \{ 10, 20, 50 \}$ are given in Table \ref{tab:Results}. Here $\alpha_{avg}$ and $\Omega_{avg}$ represent the average $\alpha$ and  $\Omega$ values, obtained across all test functions $f \in \{ \textrm{F1}, \textrm{F2}, ..., \textrm{F30} \}$.

As a baseline reference, the ``Languid'' (LPD) anakatabatic model \eqref{eq:akbLPSO} was also tested. Considering that anakatabatic inertia is a generalized form of PFIDI allowing for more complex anakatabatic models, the models hereby found through metaoptimization were all expected to yield better PSO accuracy than LPD.

\begin{table*}[!]\footnotesize
\caption{Benchmark testing of PSO with anakatabatic inertia}
\label{tab:Results}
\begin{tabular*}{\textwidth}{l@{\extracolsep{\fill}}crrrrrr}
\hline
\multicolumn{1}{l}{\multirow{2}{*}{Variant}} & \multicolumn{1}{l}{\multirow{2}{*}{Anakatabatic model}} & \multicolumn{2}{c}{$D = 10$} & \multicolumn{2}{c}{$D = 20$}  & \multicolumn{2}{c}{$D = 50$} \\
\multicolumn{2}{c}{} & \multicolumn{1}{c}{$\alpha_{avg}$} & \multicolumn{1}{c}{$\Omega_{avg}$}  & \multicolumn{1}{c}{$\alpha_{avg}$} & \multicolumn{1}{c}{$\Omega_{avg}$} & \multicolumn{1}{c}{$\alpha_{avg}$} & \multicolumn{1}{c}{$\Omega_{avg}$} \\
\hline
Standard PSO & ``Languid'' \eqref{eq:akbLPSO}  &     0.17  &    0.08  &     0.22 &     0.10 &   -0.01  &   -0.03  \\
Standard PSO & ``Flying Stork'' (Fig. \ref{fig:akbFlyingStork})  &     0.13  &   0.06  &      0.19  &      0.09  &     0.09  & 0.04  \\
Standard PSO & ``Messy Tie'' (Fig. \ref{fig:akbMessyTie})  &     -0.09 &  -0.05 &       0.11  &    0.05 &   0.09  &   0.08 \\
\hline
TVAC-PSO & ``Languid'' \eqref{eq:akbLPSO}  &     0.05  & 0.02     &      0.40  &     0.23  &    0.25 &  0.42  \\
TVAC-PSO & ``Rightward Peaks'' (Fig. \ref{fig:akbRightwardPeaks})  &     0.52  &  0.28  &  0.70 &  0.50  &  0.74 &  0.59  \\
TVAC-PSO & ``Origami Snake'' (Fig. \ref{fig:akbOrigamiSnake})  &     0.52  &    0.29  &      0.71 &     0.54 &     0.72 &   0.55  \\
\hline
\end{tabular*}
\end{table*}


The results given in Table \ref{tab:Results} allow for several comments. First of all, when comparing the $\alpha$ and $\Omega$ scores for Standard PSO it is obvious that finding an anakatabatic model which can safely outperform Languid strategy is not an easy task. The obtained models ``Flying Stork'' and ``Messy Tie'' yield contrasting results for $D = 10$, but nevertheless they are both performing worse in this category than Languid Standard PSO. On the other hand, their visual similarity may be showing in the results for $D = 50$, where they produce similar improvements in accuracy, which is also a significantly better result than the slight deterioration of accuracy of Languid PSO. However, the benchmark testing results demonstrate that, in case of Standard PSO, the first choice should still be the ``Languid'' model. The two new models may still provide better results for some optimization problems and possibly for higher problem dimensionality.

The results for TVAC-PSO anakatabatic models offer an entirely different outlook. The two obtained models (``Rightward Peaks'' and ``Origami Snake'') both produce strong and consistent improvements in method accuracy over Languid TVAC-PSO. These findings indicate that for TVAC-PSO the LPD technique could be considered obsolete, as the newly found anakatabatic models generally yield significantly greater accuracy.

\section{Conclusion}
\label{sec:Conclusion}

In our previous research we have proposed Languid Particle Dynamics (LPD) for PSO, as a Personal Fitness Improvement Dependent Inertia (PFIDI) technique. This PFIDI method makes inertia a conditional term in PSO velocity update, enabled only for particles which suceeded io improving their position in the previous iteration. As a generalization of this approach, in this paper we proposed anakatabatic inertia, which is an advanced PFIDI technique for adapting inertia weight based on a particle's fitness improvement with regard to the progress of the entire swarm.

Through metaoptimization and manual trial and arror experimentation, four anakatabatic models were found and benchmark tested. The two models designed for Standard PSO (``Flying Stork'' and ``Messy Tie'') are expected to produce improved method accuracy on a certain class of optimization problems, although on a CEC 2014 test suite (consisting of 30 test functions of vastly different types) they seem barely competitive with the ``Languid'' model (LPD). The other two models (``Rightward Peaks'' and ``Origami Snake'') were designed for TVAC-PSO and they strongly outperform ``Languid'' inertia strategy.

Quantitatively speaking, by use of the proposed anakabatic models the average accuracy of Standard PSO was improved by up to 0.09 orders of magnitude and TVAC-PSO by up to 0.59 orders of magnitude. 

Certainly, there remains a possibility that better anakatabatic models might still be found through additional or improved metaoptimization, thus continuing this line of research could prove to be even more fruitful. In particular, employing a spectrum of optimization methods of various types would surely improve the metaoptimization methodology.

Furthermore, in future research more complex anakatabatic models need to be explored, specifically model frameworks which would allow smooth $W_s(\theta)$ and $W_f(\theta)$ curves, as well as $W$-functions with discontinuities.

\section*{Software implementation}

We invite researchers and engineers to try our implementation of PSO with anakatabatic inertia by using Indago Python module. The module is free and open source, available under MIT license. It can be installed via ``pip'' command. More information on Indago is available at: \url{https://pypi.org/project/Indago/}.



%
%

\bibliography{bibliography}

\end{document}